%% file: egpaper.tex
\documentclass[10pt,twocolumn,letterpaper]{article}

\usepackage{wacv}
\usepackage{times}
\usepackage{epsfig}
\usepackage{graphicx}
\usepackage{amsmath}
\usepackage{amssymb}
\usepackage[font={small,it}]{caption}
\usepackage[export]{adjustbox}
\usepackage{multirow}
\usepackage{booktabs}
\usepackage[T1]{fontenc}
\usepackage[utf8]{inputenc}
\usepackage{authblk}

\def\wacvPaperID{0126} 

\wacvfinalcopy 

\ifwacvfinal
\usepackage[breaklinks=true,bookmarks=false]{hyperref}
\else
\usepackage[pagebackref=true,breaklinks=true,colorlinks,bookmarks=false]{hyperref}
\fi

\begin{document}

\title{Multi-stream dynamic video Summarization}


\pagestyle{empty}
\author{
Mohamed Elfeki$^{1}$, Liqiang Wang$^{2}$, and  Ali Borji\\
\vspace*{-0.35cm}
$^{1}$Microsoft, $^{2}$University of Central Florida\\
{\tt\small melfeki@microsoft.com, lwang@cs.ucf.edu, aliborji@gmail.com}
\vspace*{-0.5cm}
}

\maketitle
 \thispagestyle{empty}
\begin{abstract}
With vast amounts of video content being uploaded to the Internet every minute, video summarization becomes critical for efficient browsing, searching, and indexing of visual content. Nonetheless, the spread of social and egocentric cameras creates an abundance of sparse scenarios captured by several devices, and ultimately required to be jointly summarized. In this paper, we discuss the problem of summarizing videos recorded independently by several dynamic cameras that intermittently share the field of view. We present a robust framework that (a) identifies a diverse set of important events among moving cameras that often are not capturing the same scene, and (b) selects the most representative view(s) at each event to be included in a universal summary. Due to the lack of an applicable alternative, we collected a new multi-view egocentric dataset, Multi-Ego. Our dataset is recorded simultaneously by three cameras, covering a wide variety of real-life scenarios. The footage is annotated by multiple individuals under various summarization configurations, with a consensus analysis ensuring a reliable ground truth. We conduct extensive experiments on the compiled dataset in addition to three other standard benchmarks that show the robustness and the advantage of our approach in both supervised and unsupervised settings. Additionally, we show that our approach learns collectively from data of varied number-of-views and orthogonal to other summarization methods, deeming it scalable and generic.
\end{abstract}

\vspace{-1em}
\pagestyle{empty}
\section{Introduction}
\input{Intro}

\section{Related Work}
\input{Background}

\section{Multi-Ego: A new multi-view egocentric summarization dataset}
\input{Dataset}

\section{Approach}
\input{Method}

\section{Experiments and Results}
\input{Experiments}

\section{Conclusion}

In this work, we proposed the problem of multi-view video summarization for dynamically moving cameras that often do not share the same field-of-view. Our formulation provides the first supervised solution to multi-stream summarization in addition to an unsupervised adaptation. Unlike previous work in multi-view video summarization, we presented a generic approach that can be trained in a supervised or unsupervised setting to generate a comprehensive summary for all views with no prior assumptions on camera placement nor labels. It identifies important events across all views and selects the view(s) best illustrating each event. We also introduced a new dataset, recorded in uncontrolled environments including a variety of real-life activities. When evaluating our approach on the collected benchmark and additional three standard mutli-view benchmark datasets, our framework outperformed all baselines of state-of-the-art supervised, reinforcement and unsupervised single- and multi-view summarization methods.


{\small
\bibliographystyle{ieee_fullname}
\bibliography{egbib}
}

\end{document}

%% file: Intro.tex
\pagestyle{empty}

In a world where nearly everyone has several mobile cameras ranging from smart-phones to body-cameras, brevity becomes no longer an accessory. It is rather essential to efficiently extract relevant contents from this immense array of static and moving cameras. Video summarization aims at selecting a set of frames from a visual sequence that contains the most important and representative events. Not only is summarization useful for efficiently extracting the data substance, it also serves many other applications such as video indexing~\cite{video_indexing}, video retrieval~\cite{video_retrieval}, and anomaly detection~\cite{anomaly_detection}.

\begin{figure}
\centering
\includegraphics[width=0.52\textwidth,height=6.67cm,center]{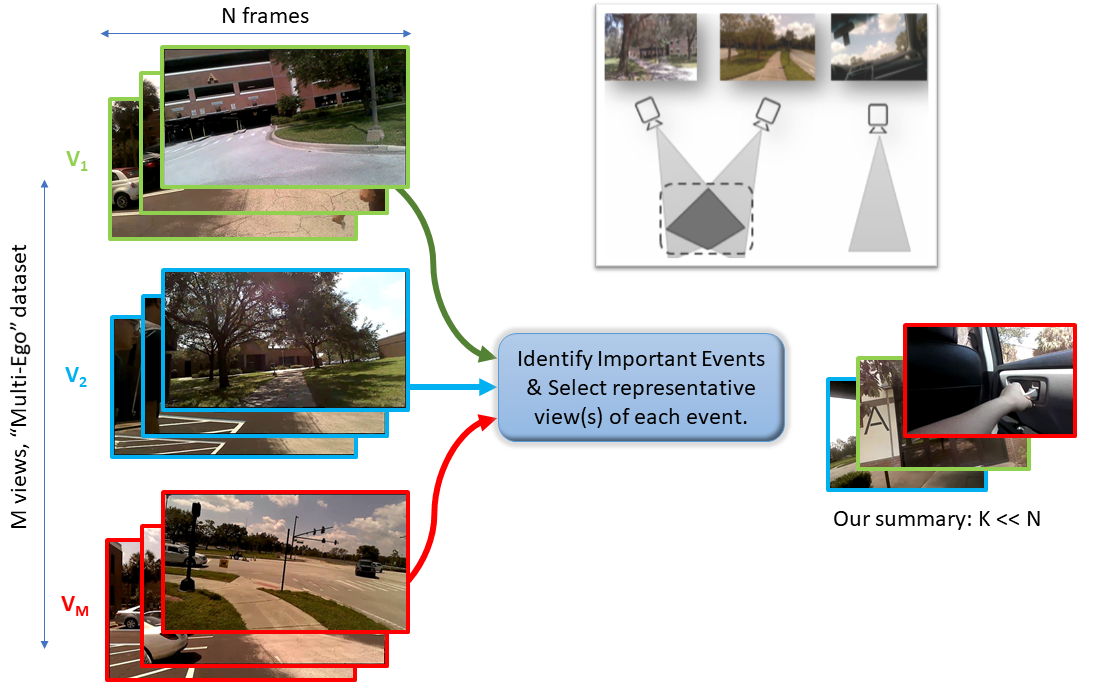}
\caption{Several views are recorded independently and intermittently overlap their fields-of-view. Our approach dynamically accounts for inter- and intra-view dependencies, providing a comprehensive summary of all views.}
\vspace{-1.67em}
\label{fig:problem_description}
\end{figure}

We consider a generic setting where multiple users record egocentric footage that is both spatially and temporally independent. Users are allowed to move freely in an uncontrolled environment. As such, cameras' fields-of-view may or may not overlap through the sequence. Unlike fixed-camera videos, egocentric footage often displays rapid changes in illumination, unpredictable camera motion, unusual composition and viewpoints, and often complex hand-object manipulations. Accordingly, the desired summary should include a diverse set of events from all viewpoints and resist the egocentric noise. Specifically, there are two types of important events to be included in the universal summary. First, events where multiple views have a substantial overlap, in which the summary include the most representative view. Second, events that are spatially independent, in which each view is processed separately from the rest.


This setting presents itself in several real-life scenarios where many egocentric videos are required to be summarized collectively. For instance, rising claims of police misconduct led to a proliferation of body cameras recordings \cite{police_2013,police_2015}. Typical police patrols contain multiple officers working 10-12 hour shifts. Although it is crucial to thoroughly inspect key details, manually going through 10-hour video content is extremely challenging and prone to human errors. Multiplying shift lengths by the number of officers on duty, it is obvious that there are copious amounts of data to analyze with no guiding index. A similar example occurs at social events such as concerts, music shows, and sports games. Those events tend to be recorded by many several cameras simultaneously that are dynamically changing their fields-of-view. Nevertheless, the final highlight summary of such events is likely to contain frames from all cameras.

Despite considerable progress in single-view video summarization for both egocentric and fixed cameras (e.g., \cite{zhang_lstm,category_specific_summarization,gong_subset_selection,adversarial_lstm}), those techniques are not readily applicable to summarizing multi-view videos. Single-view summarizers ignore the temporal order by processing simultaneously-recorded views in a sequential order to fit as a single-view input. This results in redundant and repetitive summaries that do not exhibit the multi-stream nature of the footage. On the other end of spectrum, the literature of multi-view video summarization mainly focuses on fixed surveillance camera summarization (e.g., ~\cite{panda2016video,panda2017multi}). This enables some methods to rely on geometric alignment of cameras inferring the relationship between their fields-of-view and utilizing it for a representative summary (e.g., \cite{arev2014automatic,2010_multi}). Thus, previous work mostly uses unsupervised methods that are based on heuristic-based objective functions, which are not suitable to a dynamic change in cameras' geometric positioning. A key motivation for our work is to generalize the multi-stream summarization to accommodate dynamic cameras and extend the capacity of existing supervised and unsupervised summarization techniques.

\textbf{Contributions.} We extend single-view and fixed-cameras methods to be applied on the generalized multi-stream dynamic-cameras setting. We propose a new adaptation of the widely used Determinantal Point Process (DPP)~\cite{zhang_lstm,adversarial_lstm,gong_subset_selection,cvpr_2017}, Multi-DPP, generalizes it to accommodate multi-stream setting while maintaining the temporal order. Our approach is orthogonal to other summarization approaches and can be embedded with fixed- or moving-cameras and operating on a supervised or unsupervised setting. Furthermore, our method is shown to be scalable (can be trained on labels of any available number-of-views in the supervised setting) and generic (encompasses both single-view and fixed-cameras settings as special cases). Since no existing dataset is readily applicable to evaluate such setting, we collect and annotate a new dataset, Multi-Ego. With extensive experiments, we show that our method outperforms state-of-the-art supervised and unsupervised baselines on our generic configuration as well as the special case of fixed-cameras multi-view summarization.

%% file: Background.tex
\pagestyle{empty}

\paragraph{Single-View Video Summarization}Among many approaches proposed for summarizing single-view videos supervised approaches usually stood out with best performances. In such a setting, the purpose is to simulate the patterns that people exhibit when performing the summarization task, by using human-annotated summaries. There are two-factor influence the supervised models' performance: (a) reliability of annotations, and (b) framework's modeling capability. Ensuring the reliability of annotations is evaluated based on a consensus analysis as in several benchmark datasets ~\cite{ma2002user,tvsum,gruman_important_objects}. As for the modeling capabilities, supervised approaches vary in their modeling complexity and effectiveness ~\cite{gong_subset_selection,submodular_mixture_objects,gruman_summary_transfer,gygli2014creating,rebuttal_5,rebuttal_6}. 

Recently,~\cite{rochan2018video} proposed to use convolutional sequences to summarize videos in both supervised and unsupervised settings. By formulating the problem as a sequence labeling problem, they established a connection between semantic segmentation and video summarization and used networks trained on the former to improve the latter. Others have formulated the summarization problem within a reinforcement learning paradigm either with an explicit classification reward as in ~\cite{rl_diversity} or a more subtle diversity-representativeness reward ~\cite{rl_diversity_2}. Both approaches provided relatively competitive results on single-view, nonetheless they suffer from unstable training in the multi-view setting as we detail in the experiments section.

Recurrent Neural Networks in general, and Long Short-Term Memory (LSTM)~\cite{lstm} in particular has been widely used in video processing to obtain the temporal features in videos~\cite{cap_1,cap_2,action_1,action_2}. In the recent years, using LSTMs has been a common practice to solve video summarization problem ~\cite{encoder_decoder,unsupervised_lstm_video_representation,video_highlights_recurrent_autoencoders,zhao2017hierarchical,yang2015unsupervised,li2018local,actionness}. For example, Zhang et al. \cite{zhang_lstm} use a mixture of Bi-directional LSTMs (Bi-LSTM) and Multi-Layer Perceptron to summarize single-view videos in a supervised manner. They maximize the likelihood of Determinantal point processes (DPP) measure\cite{dpp_machine_learning,dpp_mit_notes,dpp_reference} to enforce diversity within the selected summary. Also, Mahasseni et al. \cite{adversarial_lstm} present a framework that adversarially trains LSTMs, where the discriminator is used to learn a discrete similarity measure for training the recurrent encoder/decoder and the frame selector LSTMs.
\vspace{-1em}
\paragraph{Multi-view Video Summarization}Most multi-view summarization methods tend to rely on feature selection in an unsupervised optimization paradigms ~\cite{online_multi_wireless,panda2017multi,panda2016video,unpaired,multi_recent}. Fu et al. ~\cite{2010_multi} introduce the problem of multi-view video summarization as tailored for fixed surveillance cameras. They construct a spatiotemporal graph and formulate the problem as a graph-labeling task. Similarly, in \cite{panda2016video,panda2017multi,multi_recent} authors assume that cameras in a surveillance camera network have a considerable overlap in their fields-of-view. Therefore they apply well-crafted objective functions that learn an embedding space and jointly optimize for a succinct representative summary. Since those approaches target fixed surveillance cameras, they rightfully assume a significant correlation among the frames along the same view over time. In our generalized setting, cameras move dynamically and contain rapid changes in the field-of-view rendering the aforementioned assumption weak and make the problem harder to solve.


\vspace{-1em}
\paragraph{Multi-Video Summarization} Unlike multi-view, multi-video ~\cite{wang2009multi,li2010multi} focuses on spatio-temporally independent videos and thus, can be processed individually. The key challenge is scalling the framework onto a large number of input videos. \cite{ji2018hypergraph} formulated the problem into finding the dominant sets in a hypergraph. Then, refine these keyframe candidates using the web images of the same query. Recently, \cite{ji2019multi} proposed a similar method that differs in using a multi-modal weighted archetypal analysis instead of a hypergraph as a structure of the large number of web videos.

%% file: Dataset.tex
\pagestyle{empty}


While a number of multi-view datasets exist (e.g.~\cite{2010_multi,online_multi_wireless}), none of them are recorded in egocentric perspective. Therefore, we collect our own data that aligns with the established problem setting. We asked three users to independently collect a total of 12 hours of egocentric videos while performing different real-life activities. Data covers various uncontrolled environments and activities. We also ensured to present different levels of interactions among the individuals: (a) two views interacting while the third one is independent, (b) all views interacting with each other, and (c) all views independent of each other. Then, we extracted 41 different sequences that vary in length from three to seven minutes. Each sequence contains three views covering a variety of indoors and outdoors activities. We made the data more accessible for training and evaluation by grouping the sequences into 6 different collections. 

To put our dataset size (41 videos of 3-7 minutes) in perspective, we refer to the most commonly used summarization benchmarks: SumMe (25
videos of 2-4 minutes), TVSum (50 videos of 2-4 minutes)~\cite{tvsum}, Office (4 videos of 11 minutes), Lobby (3 videos of 8 minutes) and Campus (4 videos of 15 minutes)~\cite{2010_multi,online_multi_wireless}. Even though that collecting larger sizes and longer videos is desirable, nonetheless, annotating simultaneously collected views by several annotators is a notoriously hard task. In the following section, we shed some light on the difficulties encountered in that task and we propose \emph{annotating-in-stages} approach to reduce the annotation uncertainty. More details about data-collection and a behavioral analysis on the obtained annotations are provided in supplementary materials.

\subsection{Collecting User Annotations}
To annotate and process the data for the summarization task, we sub-sample the videos uniformly to one fps following ~\cite{cvpr_2017}. Then, every three consecutive frames are combined to construct a shot for an easier display to annotators. The number of frames per shot was chosen empirically to maintain a consistent activity within one shot.

We asked five human annotators to perform a three-stage annotation task. In {\emph{stage one}}, they were asked to choose the most interesting and informative shots that represent each view independently without any consideration towards the other views. To construct two-view summaries in {\emph{stage two}}, we only displayed the first two views simultaneously, while asking the users to select the shots from any of the two views that best represent both cameras. Similar to stage two, in {\emph{stage three}} the users were asked to select shots from any of the three views that best represent all the cameras. It is worth noting that the annotators were not limited to choose only one view of a certain shot, and they could choose as many as they deem important.

The \textit{annotating-in-stages} procedure explained above was employed due to the human's limited capability in keeping track of unfolding storylines along multiple views simultaneously. Consequently, using this technique resulted in a significant improvement in the consensus between user summaries compared to when we initially collected summaries in an unordered annotation task.

\subsection{Analyzing User Annotations}
To ensure the reliability and consistency of the obtained annotations, we perform a consensus analysis using two metrics: average pairwise f1-measure and selection ratio. Following~\cite{tvsum,cvpr_2017,category_specific_summarization}, we compute the average pairwise f1-measure to estimate the frame-level overlap and agreement. We calculated the f1-measure for all possible pairs of users' annotations and averaged the results across all the pairs, obtaining an average of 0.803, 0.762, and 0.834 for the first, second, and third stage respectively. 

\subsection{Creating Oracle Summaries} 
Finally, training a supervised method usually requires a single set of labels. That means in our case, we need to use only one summary per video, which is often referred to as \textit{Oracle Summary}. To create an oracle summary using multiple human-created summaries, we follow ~\cite{gong_subset_selection,doc_summarization} to greedily choose the shot that results in the largest marginal gain on the f-score, and iteratively keep repeating the greedy selection until the length of the summary reaches 15\% of the single-view length. 

%% file: Method.tex
\pagestyle{empty}


\subsection{Determinantal Point Process (DPP)}
DPP is a probabilistic measure that provides a tractable and efficient means to capture negative correlation with respect to a similarity measure~\cite{dpp_quantum,dpp_machine_learning}. Formally, a discrete point process $\mathcal{P}$ on a ground set $\mathcal{Y}$ is a probability measure on the power set $2^N$, where $N=|\mathcal{Y}|$ is the ground set size. A point process $\mathcal{P}$ is called determinantal if $\mathcal{P}(y \subseteq Y) \propto \det(L_y);\,\forall \, y \subseteq Y$. $Y$ is the selection random variable sampled according to $\mathcal{P}$ and $L$ is a symmetric semi-definite positive matrix representing the kernel. 

Kulesza et al. ~\cite{structured_dpp} proposed modeling the marginal kernel $L$ as a Gram matrix in the following manner:
\vspace{-1em}
\begin{equation}
    \mathcal{P}(y = Y) \propto \det(\Phi_y^\top \Phi_y) \prod_{i \in y} q_i^2,
    \label{eq:dpp_decomposition}
    \vspace{-1em}
\end{equation}
When optimizing the DPP kernel, this decomposition learns a ``quality score" of each item, where $q_i \geq 0$. It also allows learning a feature vector $\Phi_y$ of subset $y\subseteq\mathcal{Y}$.  In this case, the dot product $\Phi_y=[\phi_i|...|\phi_j]$, where $\phi_i^\top \phi_j \in [-1, 1]; \forall i,j\in y$ is evaluated as a ``pair-wise similarity measure" between the features of item $i, \phi_i$ and the features of item $j, \phi_j$. Thus, the DPP marginal kernel $L_y$ can be used to quantify the diversity within any subset $y$ selected from a ground set $\mathcal{Y}$. Choosing a diverse subset is equivalent to a brief representative subset since the redundancy is being minimized. Hence, it is only natural that a considerable number of document and video summarization approaches use this measure to extract representative summaries of documents and videos ~\cite{doc_summarization,adversarial_lstm,gong_subset_selection,dpp_reference}.

\subsection{Adapting DPP to Multi-stream: Multi-DPP} 
The standard DPP process described above is suitable for selecting a diverse subset from a single ground set. However, when presented with several temporally-aligned ground sets $\{\mathcal{Y}_1, \mathcal{Y}_2, ..., \mathcal{Y}_M\}$, the standard process can only be applied in one of two settings: either (a) merging all the ground sets into a single ground set $\mathcal{Y}^{merge}=\{\mathcal{Y}_1 \cup \mathcal{Y}_2 \cup ... \cup \mathcal{Y}_M\}$ and selecting a diverse subset out of the merged ground set, or (b) selecting a diverse subset from each ground set and then merging all the selected subsets ${Y}^{merge}=\{{Y}_1 \cup {Y}_2 \cup ... \cup {Y}_M\}$.

Even though that the former setting preserves the information of all elements of the ground sets, but it causes the complexity of the subset selection to exponentially grow. In practice, this leads to an accumulation of error due to overflow and underflow computations as well as substantially slower running-time. Additionally, latter setting assumes no-intersection between features of the different ground-sets. This is essentially inapplicable if the ground-sets have a significant dynamic feature overlap, leading to redundancy and compromising the very purpose of the DPP. To address these shortcomings, we propose a new adaptation of Eq. 1, called \textit{Multi-DPP}.

In Multi-DPP, ground sets are processed in parallel allowing any potential feature overlap across the ground sets to be processed temporally-appropriate and keeping a linear growth with respect to the number of streams. For every element in the ground sets, we need to represent two joint quantities: features and quality, such that they follow the following four characteristics. First, we need a model that can operate on any number of streams (i.e., generic to any number of ground sets $M$). Second, we need a joint representation of the features at each index, such that it only selects the most effective ones (i.e., invariance to noise and non-important features). Third, we need a joint representation of the qualities at each index, such that is affected by the quality of each ground set at a particular index (i.e., variance to the quality of each ground set). Forth, we need to ensure that our adaptation follows the DPP decomposition in Eq. ~\ref{eq:dpp_decomposition}, by selecting joint features $\phi_i^\top \phi_j \in [-1, 1]$, and joint qualities $q_i \geq 0; \forall i,j\in y$.

To account for joint features, we apply max-pooling choosing the most effective features across all ground sets at every index, which satisfies the feature decomposition in Eq. ~\ref{eq:dpp_decomposition}. Selecting joint qualities -on the other hand- needs to account for the quality of each ground set in every index. We use the product of all the qualities at each index. This deems the joint quality at each index to be dependent on all ground-sets while also ensuring $q^m\leq 1$. Therefore, we generalize the Determinantal Point Process based on the decomposition in Eq. ~\ref{eq:dpp_decomposition} as follows:
\vspace{-1em}
\begin{equation}
\begin{split}
\mathcal{P}(Y = y) \propto \det(\Phi_y^\top \Phi_y) \prod_{m=1}^{M}\prod_{i \in y_m} [q_i^m]^2\\
\phi_j = max(\phi_j^1, ..., \phi_j^M)\,\,; \forall j\in y\,\,\,\,\,\,\,\,\,\,\,\,
\end{split}
\label{eq:multi_dpp}
\end{equation}
where $M$ is the number of the ground sets and $y_m$ is the subset selected from ground set $m$. This decomposition allows both a scalable multi-stream (by constructing a joint feature representation with max-pooling), and monitoring the egocentric-introduced noise (by learning an independent quality measure for each view at each time-step).

\paragraph{Summarizing videos using Multi-DPP.}Since Multi-DPP formulation of Eq.~\ref{eq:multi_dpp} does not require any extra supervisory signals, it can be adopted to an optimization formula for both supervised and unsupervised training. In particular, we follow ~\cite{dpp_machine_learning} in defining the similarity measure of supervised summarization approaches based on a Maximum Likelihood Estimation of the Multi-DPP measure with respect to the ground-truth labels as follows:
\begin{equation}
    \theta^* = argmax_{\theta} \sum_i log\big\{P(Y^{(i)}=y^{(i)*}; L^{(i)}(\theta)\big\}
    \label{eq:mle_multi_dpp}
    \vspace{-1em}
\end{equation}
where $\theta$ is the set of supervised parameters, $y^*$ is the target subset (i.e., ground-truth) and $i$ indexes training examples.

For unsupervised summarization, we define the Multi-DPP loss based on a diversity regularization introduced in ~\cite{adversarial_lstm} that aims to only increase diversity since no summary labels are being provided.
\begin{equation}
    \theta^* = argmax_{\theta}\,\, log\big\{P(Y; L^{(i)}(\theta)\big\}
    \label{eq:map_multi_dpp}
\end{equation}
where $\theta$ is the set of unsupervised parameters. 

Finally we note that our supervised and unsupervised adaptations are orthogonal to other summarization approaches and can be embedded to allow any DPP-based approach (e.g.,~\cite{zhang_lstm,adversarial_lstm,bmvc,sharghi2018improving,elfeki2018gdpp}) to summarize multi-stream data while preserving the temporal order and monitoring the quality of a dynamic input. Additionally, Multi-DPP is equivalent to the standard DPP decomposition in Eq.1 when $M=1$ at Eq.2. This renders Multi-DPP summarization approach as a generalization of the standard single-view summarization DPP approaches as well as orthogonal to other summarization approaches that allows them to process multi-stream data in a proper temporal order. The discussed theoretical advantage of such generalization will be further analyzed empirically at Section 5.3.

\begin{figure}[t]
\centering
\includegraphics[width=0.48\textwidth,height=4.8cm, center]{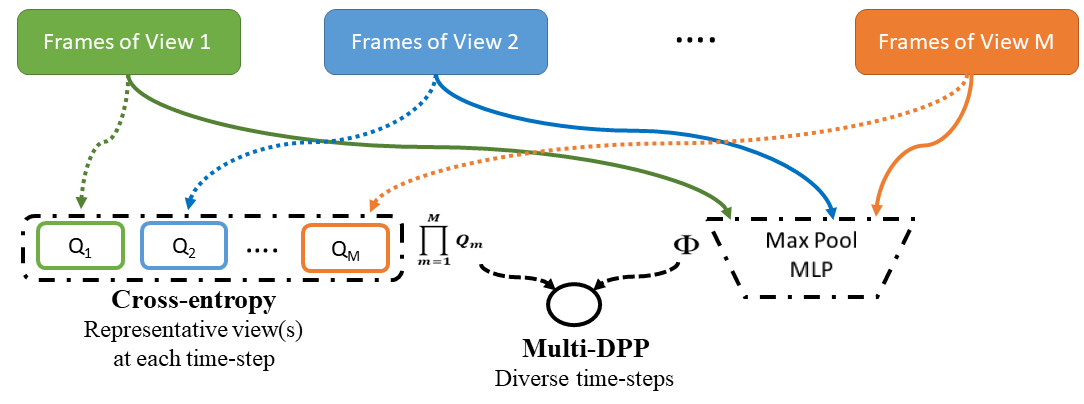}
\caption{Multi-DPP is applied to increase diversity within the selected time-steps. When view labels are available, we also use cross-entropy to learn representative view(s) at each time-step.}
\vspace{-1.27em}
\label{fig:system_description}
\end{figure}

\subsection{Summarization Framework}
Figure ~\ref{fig:system_description} shows the input as $M$ independent views, with $N$ frames at each view. We follow ~\cite{zhang_lstm,adversarial_lstm,bmvc,sharghi2018improving} in constructing features of each frame across the streams. First, spatial features are extracted from each frame at each view using a pre-trained CNN. Then, spatial features are temporally processed using a Bidirectional LSTM layer. By aggregating both spatial and temporal features, we obtain a comprehensive spatio-temporal feature vector of each frame at each view. We choose to share the weights of the Bi-LSTM layer across the views for two reasons: (a) it allows the system to operate on any number of views without increasing the number of trainable parameters which alleviates overfitting, and (b) learning temporal features is independent of the view, thus it utilizes data from all views to produce better temporal modeling.


We break down our objective into two tasks: selecting diverse events and identifying the view(s) contributing to illustrating each selected event in summary. In first task, to select diverse events, we construct a feature set accounting for all the views at each time-step. We do so by max-pooling the spatio-temporal features from all the views, resulting in the most prominent feature at each index of the feature vector. We follow max-pooling by a two-layer Multi-Layer Perceptron (MLP) that applies non-linear activation on joint features that are represented as $\Phi$ in Eq. ~\ref{eq:multi_dpp}.

The second task, however, is used to identify the most representative view(s) at each event. We use a two-layer MLP that classifies each view at each time step. Formulating this task as a classification problem serves three purposes. First, it selects the views that are included in the summary, which is an intrinsic part of the solution. Second, it regularizes the process of learning the importance of each event by not selecting any view when the time-step is non-important. Finally, the classification confidence of view $m$ can be used to represent the quality ($q_n^m$) at time-step $n$. This is later used to compute the Multi-DPP measure that determines which time-steps are selected. In the case of non-overlapping views, the framework may need to select multiple views at the same time-step. That's why, we conduct an independent view classification by applying binary classification, which allows classifying each view independently from the rest.

Similar to the weights of the Bi-LSTM, the view classifier MLP weights are also shared across the views for two reasons. First, it uses the same number of trainable parameters for any number-of-views data, resulting in fewer trainable parameters which limit the problem of overfitting to training data. Second, it establishes a view-dependent classification. That is, at any time-step, choosing a representative view among all the views is affected by the relative quality of all the views, rather than each one independently. During training, we start by estimating the quality $q_n^m$ of each view $m$ at each time-step $n$, which serves as the view selection. Then we evaluate Multi-DPP measure by merging the computed $q_n^m$ with the joint-features $\Phi$ as in Eq.~\ref{eq:multi_dpp}.

In our supervised setting, we optimize the view(s) selection procedure by using the binary cross-entropy objective: $-\frac{1}{M} \sum_{m=1}^{M} \sum_{n=1}^{N} y_n^m log(p_n^m)$; where $y_n^m, p_n^m$ are the ground truth and model's prediction for the time-step $n$ in view $m$. We jointly optimize the framework by minimizing the sum of the cross entropy loss as well as Eq.~\ref{eq:mle_multi_dpp} and using the Oracle summary as the ground-truth in the supervised setting. In the unsupervised setting, view selection weights are only learned by learning the quality $q_n^m$ from the Multi-DPP measure and we only optimize the Multi-DPP loss criterion Eq.~\ref{eq:map_multi_dpp}.

Lastly, while input views are not required to be temporally aligned, they are assumed to have timestamps. This is a commonly held assumption in previous multi-view literature (e.g.,~\cite{2010_multi,multi_1}) due to its default presence in nearly all modern recording devices. If given non-aligned views, our framework can process any number of views at each time-step since the weights of the Bi-LSTM and the MLPs are shared among the views.

\subsection{Multi-view supervised scalability}

Supervised summarization tends to have a superior generalization performance when compared to unsupervised ones, e.g.,~\cite{gong_subset_selection,category_specific_summarization,zhang_lstm,adversarial_lstm}. Relying on human-annotated labels allows learning generic behavioral patterns instead of customized heuristics as in most unsupervised approaches. Nonetheless, supervision requires an abundance of labeled training data. Thus, a crucial concern of a multi-view supervised system is to be scalable in order to utilize all available forms of labels for an improved performance. Obviously, unsupervised systems do not undergo this challenge since they do not utilize labels. 

In particular, a scalable multi-view video summarizer is invariant to view order and number-of-views, and therefore can learn from any data regardless of those properties. First, invariance to view order implies producing the same summary for input views $(v_i, v_j, v_k)$ as to $(v_j, v_i, v_k); \forall i,j,k \in \{1, 2, .., M\}$, for all possible permutations of $(i,j,k)$. Our approach satisfies this requirement by constructing joint-features via max-pooling. Thus, summary is only shaped by the most effective features regardless of view order.

The second condition, invariance to number-of-views, entails the ability to train on data with \textit{varying} numbers-of-views and test on data of \textit{any} number-of-views. Satisfying this condition requires the number of trainable parameters to be invariant from the number-of-views of the input. This way the same set of parameters can be used to train/test on data with any number-of-views. We followed two techniques ensuring a fixed number of trainable parameters: (a) max pooling view-specific features, and (b) weight-sharing for Bi-LSTM and view selection layers. Firstly, Applying max-pooling on view-specific features produces a fixed-size joint feature vector that is invariant from the number-of-views in the input. Additionally, choosing the prominent features across views entails learning intra-view dependencies. Secondly, weight sharing across Bi-LSTM view-streams and view selection layers ensures our framework has a single set of trainable parameters for each of those layers regardless number-of-views. 

%% file: Experiments.tex
\pagestyle{empty}


\begin{table*}[t]
\centering
  \begin{adjustbox}{max width=\textwidth}
\begin{tabular}{@{}c|cccccccc@{}}
\toprule
                                                                                                          &                                                                                                                      & \multicolumn{3}{c}{Two-View}               &  & \multicolumn{3}{c}{Three-View}               \\
                                                                                                          &                                                                                                                      & Precision      & Recall         & F1-Score       &  & Precision      & Recall         & F1-Score       \\ \midrule
Random Baseline                                                                                            & Uniform Sampling                                                                                        & 9.83           & 10.65          & 9.85           &  & 5.83           & 5.16           & 5.77           \\ \midrule
\multirow{2}{*}{\begin{tabular}[c]{@{}c@{}}{Unsupervised}\\ \& Sub-modular \\ Multi-View\end{tabular}}                         & feature selection~\cite{nie2010efficient}                                                                   & 17.83          & 19.15          & 17.46          &  & 12.33          & 16.28          & 10.70           \\
                                                                                                          & joint embedding~\cite{panda2017multi}                                                                              & 18.37          & 25.20          & 20.66          &  & 13.88          & 24.85          & 17.17          \\
                                                                                                          & Unpaired Data~\cite{unpaired}                                                                            & 21.26          & 22.16          & 21.81          &  & 19.62          & 19.93          & 19.41          \\
                                                                                                          & Sub-modular~\cite{submodular_mixture_objects}                                                                              & 19.91          & 25.21          & 22.71          &  & 18.49          & 22.71          & 20.19          \\  \midrule
\multirow{2}{*}{\begin{tabular}[c]{@{}c@{}}\\Unsupervised\\ Single-View \end{tabular}} & Adversarial~\cite{adversarial_lstm}: Merge-Views                                                                                                         & 21.16          & 23.42          & 22.35          &  & 20.2          & 18.94          & 19.76          \\
                                                                                                          & Adversarial~\cite{adversarial_lstm}: Merge-Summaries                                                                                                     & 20.61          & 22.05          & 21.12          &  & 19.32          & 18.24          & 18.96          \\  & Convolutional~\cite{rochan2018video}: Merge-Views                                                                                                         & 21.05          & 22.92          & 22.26          &  & 19.86          & 20.68          & 20.13          \\
                                                                                                          & Convolutional~\cite{rochan2018video}: Merge-Summaries                                                                                                     & 20.64          & 22.34          & 21.87          &  & 16.52          & 20.47          & 18.91          \\ \midrule
\textbf{Ours-unsupervised}                           & \textbf{Multi-DPP}                                                                                                       & \textbf{23.91} & \textbf{24.72} & \textbf{24.18} &  & \textbf{21.96} & \textbf{22.24} & \textbf{22.61}  \\  \midrule  \midrule
\multirow{2}{*}{\begin{tabular}[c]{@{}c@{}}\\Supervised\\\& RL\\ Single-View \end{tabular}} & LSTM~\cite{zhang_lstm}: Merge-Views                                                                                                         & 27.87          & 28.57          & 27.67          &  & 23.25          & 23.87          & 22.95          \\
                                                                                                          & LSTM~\cite{zhang_lstm}: Merge-Summaries                                                                                                     & 26.61          & 27.25          & 26.43          &  & 22.86          & 23.59          & 22.76          \\
                                                                                                          & Convolutional~\cite{rochan2018video}: Merge-Views                                                                                                     & 26.84          & 26.01          & 26.38          &  & 22.28          & 23.47          & 22.92          \\  & RL Diversity~\cite{rl_diversity}: Merge-Summaries                                                                                                         & 25.02          & 27.00          & 25.97          &  & 23.78          & 22.14          & 23.14          \\
                                                                                                          & RL Classification~\cite{rl_diversity_2}: Merge-Summaries                                                                                                     & 26.01          & 26.71          & 26.27          &  & 22.74          & 23.68          & 23.37          \\  \midrule
\multirow{2}{*}{\begin{tabular}[c]{@{}c@{}}(Ablation Study) \\ \textbf{Ours-supervised} \\ { }\end{tabular}}                             & Only Cross-Entropy (CE)                                                                                                        & 27.33 & 27.83 & 27.13 &  & 21.33 & 22.03 & 21.10  \\            &
\textbf{Full: Multi-DPP + CE}                                                                                                        & \textbf{28.58} & \textbf{29.05} & \textbf{28.30} &  & \textbf{25.06} & \textbf{25.79} & \textbf{25.03}  \\    \bottomrule
\end{tabular}
\end{adjustbox}
\caption{\textit{MultiEgo} benchmarking for two-view and three-view settings. Ours consistently outperforms the baselines on all the measures. We also run an ablation study to show the effect of optimizing the supervised Multi-DPP measure as compared to using only Cross-Entropy.}
\label{tab:evaluation}
\vspace{-1.37em}
\end{table*}

\subsection{Baseline Methods}
Since our supervised approach is the first supervised multi-view summarization method, we could not compare with other supervised Multi-View approaches. Nonetheless, we compare our criterion with supervised and unsupervised single-view, and unsupervised multi-view summarizations. Additionally, we include Reinforcement Learning baselines that showed competitive performance on single-view videos.

To apply the single-view configuration on multi-view videos, we examine two settings:
\begin{itemize}
\item \textit{Merge-Views}: Aggregating views then summarizing aggregate footage using a single-view summarizer. Summary is consistent if the views are independent.
\item \textit{Merge-Summaries}: Summarizing each view independently and then aggregating the summaries. Complementary to the former setting, this should result in a consistent summary if the summaries are independent.
\end{itemize}

In our experiments, we observed that the supervised version of Convolutional Sequences~\cite{rochan2018video} tends to diverge when using Merge-summaries method in training due to relatively short videos in their case. Thus, we compare with the more reliable version of Merge-views. On the contrary, reinforcement learning methods ~\cite{rl_diversity,rl_diversity_2} tend to be unstable for the merge-views due to the long sequential input where the reward is usually far away from the start of the sequence, and thus it may lead to vanishing the gradients. So, we compare with the merge-summary concatenation, where the reward function tends to be more stable. \emph{This observed instability faced in training the baselines establishes a better motive for developing an objective like ours that is curated to be independent of number views, making it tractable during training/testing when the number of views is large, and at the same time incorporates the information from all views while preserving temporal ordering.}

\subsection{Experimental Setup}
We use GoogLeNet ~\cite{google_net} features for all the methods as an input. For a fair comparison, we train all supervised baselines ~\cite{submodular_mixture_objects,zhang_lstm} and Ours with the same experimental setup: iterations number, batch size, and optimization. We note that all neural-network models have the same architecture (same number of trainable parameters) and only differ in the objective function and their training strategy to ensure a fair comparison.

The supervised frameworks are trained for twenty iterations with a batch size of 10 sequences. Adam optimizer is used to optimize the losses with a learning rate of 0.001. After each iteration, we calculate the mean validation loss and only evaluate the model with the best validation loss across all iterations. We discuss further details of the architecture and training in the supplementary materials. 

As discussed in section 3.1, we categorize our dataset sequences into six collections to facilitate the training and evaluation. In our experiments, we follow a round-robin approach to train-validate-test the supervised/semi-supervised learning frameworks. We use four collections for training, one for validation, and one for testing across all the 30 different combinations of collections. Since no training is required for unsupervised approaches, we only test methods on each collection separately and report their means.

To evaluate the summaries produced by all the methods, we follow the protocols in ~\cite{adversarial_lstm,zhang_lstm,encoder_decoder,tvsum} to compare the predictions against the oracle summary. We start by temporally segmenting all views using the KTS algorithm ~\cite{category_specific_summarization} to non-overlapping intervals. Then, we repetitively extract key-shot based summaries using MAP ~\cite{gruman_summary_transfer} while setting the threshold of summary length to be 15\% of a single view's length. For each of the selected shots, we consider all of its frames to be included in the summary.

\subsection{Performance Evaluation}
We follow~\cite{panda2017diversity,panda2017multi,zhang_lstm,adversarial_lstm,2010_multi} in using f1-score, precision, and recall to evaluate the quality of the produced summaries by comparing frame-level correspondences between the predicted summary and the ground-truth summary. Table ~\ref{tab:evaluation} shows the mean precision, recall, and F1-score across all the combinations of training-validation-testing for both the two-view setting and three-view setting. 

In general, supervised frameworks perform better than unsupervised ones due to learning from human annotations. For unsupervised methods, ~\cite{panda2017multi,nie2010efficient,submodular_mixture_objects,unpaired} obtain the lowest performance indicating their inability to adapt to visual changes occurring in egocentric motion due to the lack of summary labels. However, using adversarial training~\cite{adversarial_lstm} seems to improve the results even with a single-view setting since the learning distribution converges to true data distribution, and it better learns to isolate egocentric-noise. Similarly, the supervised single-view BiLSTM ~\cite{zhang_lstm} and Convolutional Sequences~\cite{rochan2018video} reasonably adapt to egocentric visual noise utilizing the summary labels. Only our model monitors the egocentric-introduced noise and process data in a proper temporal order, achieving the best performance in both unsupervised and supervised comparisons.

To study the impact of enforcing diversity, we run an ablation study by evaluating our supervised approach with only optimizing cross-entropy loss(Ours: Cross-Entropy (CE) in Table~\ref{tab:evaluation}). This corresponds to training our model by only selecting representative views, without explicitly enforcing diversity. Evidently, adding Multi-DPP measure to the CE loss improves the results, especially in the three-view setting due to the increase of input footage required to diversify. It is worth noting that using only Multi-DPP is equivalent to our unsupervised version. 

Generally, it can be noticed that performance in the two-view setting is higher than that in the three-view setting, although methods' ranking remains the same. This is because of the increase in problem complexity when considering more views to be summarized, causing the performance to drop. Additionally, the performance gap increases as we move from two-view to three-view setting. Theoretically, we expect approaches such as ~\cite{rochan2018video,zhang_lstm,adversarial_lstm,rl_diversity} drop performance as the number of views grows and this is backed up empirically. Secondly, whether we concatenate views or concatenate summaries in order to adapt ~\cite{zhang_lstm,rl_diversity,rochan2018video,adversarial_lstm}, the complexity of the adaptation is unnecessarily high (either a larger DPP kernel in case of view concatenation and processing each view separately in summary concatenation scenario). Our proposed approach uses a maxpool operation as well as view quality multiplication to represent all views while preserving computational/memory efficiency.

\begin{table}[t]
\centering
\begin{adjustbox}{max width=0.9\textwidth}
\begin{tabular}{@{}cccc@{}}
\toprule
Method & Office & Campus & Lobby \\ \midrule
\begin{tabular}[c]{@{}c@{}}{Graph \cite{peng2006clip}}\end{tabular} & 41.3 & 49.1 & 73.4 \\
RandomWalk \cite{2010_multi} & 75.8 & 61.6 & 86.8 \\
RoughSets \cite{multi_2} & 75.8 & 62.1 & 84.2 \\
BipartiteOPF \cite{multi_1} & 81.8 & 71.8 & 88.2 \\
Unpaired Data~\cite{unpaired}  & 91.0 & 80.5 & 89.3 \\
Joint embedding \cite{panda2017multi} & 89.4 & 77.8 & 92.5 \\
Convolutional~\cite{rochan2018video}-Unsup & 90.2 & 78.6 & 92.5 \\
Convolutional~\cite{rochan2018video}-Sup & 94.0 & 81.9 & 93.0 \\
RL Diversity~\cite{rl_diversity} & 92.9 & 80.6 & 91.4 \\
RL Classification~\cite{rl_diversity_2} & 92.1 & 82.5 & 92.2 \\
Ours-unsupervised & 90.7 & 81.2 & 92.7 \\
Ours-supervised & \bf 94.2 & \bf 86.1 & \bf 93.4 \\ \bottomrule
\end{tabular}
\end{adjustbox}
\captionof{table}{Fixed-cameras multi-view f1-scores. We train our supervised model on \textbf{Multi-Ego} and test it on three datasets.}
\label{tab:multi_view}
\vspace{-1.5em}
\end{table}



Finally, we investigate the performance of our approach on fixed-cameras multi-view setting, which is a special case of our generic configuration. We evaluate our model on three standard fixed-cameras multi-view benchmarks: Office, Campus, and Lobby datasets~\cite{2010_multi,online_multi_wireless}. We train our supervised model on our \emph{Multi-Ego} dataset, and evaluate it on the testing dataset. Table \ref{tab:multi_view} shows a substantial success in transferring the learning from one domain (egocentric multi-view) to another domain (static multi-view) without the need to specifically-tailored training data. Thus, we provide the first supervised multi-view summarization that significantly outperforms state-of-the-art unsupervised approaches while only being trained on our data. Additionally, our unsupervised model outperforms them due to explicitly enforcing diversity and quality constraint. The consistent advantage in the three experimental environments for both our supervised and unsupervised models demonstrates the versatility of the proposed approach in handling static/egocentric videos in a generic summarization setting.

\begin{table}[t]
\centering
\begin{adjustbox}{max width=0.9\textwidth}
\begin{tabular}{@{}cc|ccc@{}}
\toprule
Test                        & Train                                                               & Precision & Recall & F1-Score \\ \midrule
\parbox[t]{2mm}{\multirow{3}{*}{\rotatebox[origin=c]{90}{two-view\,\,\,\,}}}   & 2$\times$two-view                                                         &  29.83     & 29.77  & 29.67 \\
                            &  3$\times$three-view    &  29.77     & 30.30  & 30.2  \\  \cmidrule(l){2-5}
                            & \begin{tabular}[c]{@{}c@{}}2$\times$two-view +\\3$\times$three-view\end{tabular} &  \bf 34.37     & \bf 35.03  & \bf 34.33    \\ \midrule  \midrule
\parbox[t]{2mm}{\multirow{3}{*}{\rotatebox[origin=c]{90}{three-view\,\,\,}}}   & 2$\times$three-view                                                         &  18.53     & 18.80  & 18.33 \\
                            &  2$\times$two-view    &  18.23     & 18.27  & 17.67  \\   \cmidrule(l){2-5}
                            & \begin{tabular}[c]{@{}c@{}}3$\times$two-view +\\2$\times$three-view\end{tabular} & \bf  21.53     & \bf 21.87  & \bf 21.33  \\ \bottomrule
\end{tabular}%
\end{adjustbox}
\captionof{table}{Scalability Analysis: Our framework can be trained and tested on data of different number-of-views.}
\label{tab:scalability}

\vspace{-1.67em}
\end{table}

\subsection{Supervised Scalability Analysis}
In this section, we study our supervised framework's capability to learn from a varying number-of-views in a sequence by verifying if the training process can exploit any increase in data regardless of its numbers-of-views. We start by splitting our data into two categories of nearly the same number of sequences: (a) three-view (Collections: Indoors-Outdoors, SeaWorld, Supermarket), and (b) two-view (Collections: Car-Ride, College-Tour, Library). We investigate the performance of three train/test configurations where testing data is limited to a single category:

\noindent\textit{1. Same category training (2$\times$two-view\& 1$\times$two-view):} Train on 2 collections from same category as testing.\\
\noindent\textit{2. Different category training (3$\times$two-view\& 3$\times$three-view):} Train on 3 collections from one category, and then test it on a collection belonging to a different category.\\
\noindent\textit{3. Training using Data from the two categories (3$\times$two-view + 2$\times$two-view\& 2$\times$two-view + 3$\times$two-view):} Train on data from different categories, and test it on a collection from one of the categories in the training data.


As shown in Table~\ref{tab:scalability}, training our framework on same categories or different categories obtain comparable results when testing on both two-view and three-view settings. However, increasing training data size by combining both categories significantly improves the results. This shows that our model can be trained and tested on data of various number-of-views and also is able take advantage of any data increase with no regard to its number-of-views setting.

%% file: egpaper.bbl
\begin{thebibliography}{10}\itemsep=-1pt

\bibitem{arev2014automatic}
Ido Arev, Hyun~Soo Park, Yaser Sheikh, Jessica Hodgins, and Ariel Shamir.
\newblock Automatic editing of footage from multiple social cameras.
\newblock {\em ACM Transactions on Graphics (TOG)}, 33(4):81, 2014.

\bibitem{police_2015}
Barak Ariel, William~A Farrar, and Alex Sutherland.
\newblock The effect of police body-worn cameras on use of force and
  citizens’ complaints against the police: A randomized controlled trial.
\newblock {\em Journal of quantitative criminology}, 31(3):509--535, 2015.

\bibitem{bmvc}
Chen.
\newblock Video to text summary: Joint video summarization and captioning with
  recurrent neural networks.
\newblock In {\em BMVC}, pages 1--10, 2017.

\bibitem{actionness}
Mohamed Elfeki and Ali Borji.
\newblock Video summarization via actionness ranking.
\newblock {\em Winter Applications in Computer Vision (WACV)}, 2019.

\bibitem{elfeki2018gdpp}
Mohamed Elfeki, Camille Couprie, Morgane Riviere, and Mohamed Elhoseiny.
\newblock Gdpp: Learning diverse generations using determinantal point process.
\newblock {\em arXiv preprint arXiv:1812.00068}, 2018.

\bibitem{rebuttal_6}
Chenyou Fan, Jangwon Lee, Mingze Xu, Krishna~Kumar Singh, Yong~Jae Lee, David~J
  Crandall, and Michael~S Ryoo.
\newblock Identifying first-person camera wearers in third-person videos.
\newblock {\em arXiv preprint arXiv:1704.06340}, 2017.

\bibitem{anomaly_detection}
Yachuang Feng, Yuan Yuan, and Xiaoqiang Lu.
\newblock Learning deep event models for crowd anomaly detection.
\newblock {\em Neurocomputing}, 219:548--556, 2017.

\bibitem{2010_multi}
Yanwei Fu, Yanwen Guo, Yanshu Zhu, Feng Liu, Chuanming Song, and Zhi-Hua Zhou.
\newblock Multi-view video summarization.
\newblock {\em IEEE Transactions on Multimedia}, 12(7):717--729, 2010.

\bibitem{gong_subset_selection}
Boqing Gong, Wei-Lun Chao, Kristen Grauman, and Fei Sha.
\newblock Diverse sequential subset selection for supervised video
  summarization.
\newblock In Z. Ghahramani, M. Welling, C. Cortes, N.~D. Lawrence, and K.~Q.
  Weinberger, editors, {\em Advances in Neural Information Processing Systems
  27}, pages 2069--2077. Curran Associates, Inc., 2014.

\bibitem{dpp_mit_notes}
Swati Gupta.
\newblock 1 determinantal point processes.

\bibitem{gygli2014creating}
Michael Gygli, Helmut Grabner, Hayko Riemenschneider, and Luc Van~Gool.
\newblock Creating summaries from user videos.
\newblock In {\em European conference on computer vision}, pages 505--520.
  Springer, 2014.

\bibitem{submodular_mixture_objects}
Michael Gygli, Helmut Grabner, and Luc Van~Gool.
\newblock Video summarization by learning submodular mixtures of objectives.
\newblock In {\em Proceedings of the IEEE Conference on Computer Vision and
  Pattern Recognition}, pages 3090--3098, 2015.

\bibitem{lstm}
Sepp Hochreiter and J{\"u}rgen Schmidhuber.
\newblock Long short-term memory.
\newblock {\em Neural computation}, 9(8):1735--1780, 1997.

\bibitem{video_indexing}
Richang Hong, Lei Li, Junjie Cai, Dapeng Tao, Meng Wang, and Qi Tian.
\newblock Coherent semantic-visual indexing for large-scale image retrieval in
  the cloud.
\newblock {\em IEEE Transactions on Image Processing}, 2017.

\bibitem{encoder_decoder}
Zhong Ji, Kailin Xiong, Yanwei Pang, and Xuelong Li.
\newblock Video summarization with attention-based encoder-decoder networks.
\newblock {\em arXiv preprint arXiv:1708.09545}, 2017.

\bibitem{ji2018hypergraph}
Zhong Ji, Yuanyuan Zhang, Yanwei Pang, and Xuelong Li.
\newblock Hypergraph dominant set based multi-video summarization.
\newblock {\em Signal Processing}, 148:114--123, 2018.

\bibitem{ji2019multi}
Zhong Ji, Yuanyuan Zhang, Yanwei Pang, Xuelong Li, and Jing Pan.
\newblock Multi-video summarization with query-dependent weighted archetypal
  analysis.
\newblock {\em Neurocomputing}, 332:406--416, 2019.

\bibitem{multi_1}
Sanjay~K Kuanar, Kunal~B Ranga, and Ananda~S Chowdhury.
\newblock Multi-view video summarization using bipartite matching constrained
  optimum-path forest clustering.
\newblock {\em IEEE Transactions on Multimedia}, 17(8):1166--1173, 2015.

\bibitem{structured_dpp}
Alex Kulesza and Ben Taskar.
\newblock Structured determinantal point processes.
\newblock In {\em NIPS}, 2010.

\bibitem{doc_summarization}
Alex Kulesza and Ben Taskar.
\newblock Learning determinantal point processes.
\newblock 2011.

\bibitem{dpp_machine_learning}
Alex Kulesza, Ben Taskar, et~al.
\newblock Determinantal point processes for machine learning.
\newblock {\em Foundations and Trends{\textregistered} in Machine Learning},
  5(2--3):123--286, 2012.

\bibitem{gruman_important_objects}
Yong~Jae Lee and Kristen Grauman.
\newblock Predicting important objects for egocentric video summarization.
\newblock {\em International Journal of Computer Vision}, 114(1):38--55, 2015.

\bibitem{multi_2}
Ping Li, Yanwen Guo, and Hanqiu Sun.
\newblock Multi-keyframe abstraction from videos.
\newblock In {\em 2011 18th IEEE International Conference on Image Processing},
  pages 2473--2476. IEEE, 2011.

\bibitem{li2010multi}
Yingbo Li and Bernard Merialdo.
\newblock Multi-video summarization based on video-mmr.
\newblock In {\em 11th International Workshop on Image Analysis for Multimedia
  Interactive Services WIAMIS 10}, pages 1--4. IEEE, 2010.

\bibitem{li2018local}
Yandong Li, Liqiang Wang, Tianbao Yang, and Boqing Gong.
\newblock How local is the local diversity? reinforcing sequential
  determinantal point processes with dynamic ground sets for supervised video
  summarization.
\newblock In {\em Proceedings of the European Conference on Computer Vision
  (ECCV)}, pages 151--167, 2018.

\bibitem{action_2}
Jun Liu, Amir Shahroudy, Dong Xu, and Gang Wang.
\newblock Spatio-temporal lstm with trust gates for 3d human action
  recognition.
\newblock In {\em European Conference on Computer Vision}, pages 816--833.
  Springer, 2016.

\bibitem{ma2002user}
Yu-Fei Ma, Lie Lu, Hong-Jiang Zhang, and Mingjing Li.
\newblock A user attention model for video summarization.
\newblock In {\em Proceedings of the tenth ACM international conference on
  Multimedia}, pages 533--542. ACM, 2002.

\bibitem{dpp_quantum}
Odile Macchi.
\newblock The coincidence approach to stochastic point processes.
\newblock {\em Advances in Applied Probability}, 7(1):83--122, 1975.

\bibitem{adversarial_lstm}
Behrooz Mahasseni, Michael Lam, and Sinisa Todorovic.
\newblock Unsupervised video summarization with adversarial lstm networks.
\newblock In {\em Proc. IEEE Conf. Comput. Vis. Pattern Recognit}, pages 1--10,
  2017.

\bibitem{multi_recent}
Jingjing Meng, Suchen Wang, Hongxing Wang, Junsong Yuan, and Yap-Peng Tan.
\newblock Video summarization via multi-view representative selection.
\newblock In {\em Proceedings of the IEEE International Conference on Computer
  Vision Workshops}, pages 1189--1198, 2017.

\bibitem{nie2010efficient}
Feiping Nie, Heng Huang, Xiao Cai, and Chris~H Ding.
\newblock Efficient and robust feature selection via joint l2, 1-norms
  minimization.
\newblock In {\em Advances in neural information processing systems}, pages
  1813--1821, 2010.

\bibitem{online_multi_wireless}
Shun-Hsing Ou, Chia-Han Lee, V~Srinivasa Somayazulu, Yen-Kuang Chen, and
  Shao-Yi Chien.
\newblock On-line multi-view video summarization for wireless video sensor
  network.
\newblock {\em IEEE Journal of Selected Topics in Signal Processing},
  9(1):165--179, 2015.

\bibitem{cap_2}
Pingbo Pan, Zhongwen Xu, Yi Yang, Fei Wu, and Yueting Zhuang.
\newblock Hierarchical recurrent neural encoder for video representation with
  application to captioning.
\newblock In {\em Proceedings of the IEEE Conference on Computer Vision and
  Pattern Recognition}, pages 1029--1038, 2016.

\bibitem{panda2017multi}
Rameswar Panda and Amit~Roy Chowdhury.
\newblock Multi-view surveillance video summarization via joint embedding and
  sparse optimization.
\newblock {\em IEEE Transactions on Multimedia}, 2017.

\bibitem{panda2016video}
Rameswar Panda, Abir Dasy, and Amit~K Roy-Chowdhury.
\newblock Video summarization in a multi-view camera network.
\newblock In {\em Pattern Recognition (ICPR), 2016 23rd International
  Conference on}, pages 2971--2976. IEEE, 2016.

\bibitem{panda2017diversity}
Rameswar Panda, Niluthpol~Chowdhury Mithun, and Amit Roy-Chowdhury.
\newblock Diversity-aware multi-video summarization.
\newblock {\em IEEE Transactions on Image Processing}, 2017.

\bibitem{peng2006clip}
Yuxin Peng and Chong-Wah Ngo.
\newblock Clip-based similarity measure for query-dependent clip retrieval and
  video summarization.
\newblock {\em IEEE Transactions on Circuits and Systems for Video Technology},
  16(5):612--627, 2006.

\bibitem{category_specific_summarization}
Danila Potapov, Matthijs Douze, Zaid Harchaoui, and Cordelia Schmid.
\newblock Category-specific video summarization.
\newblock In {\em European conference on computer vision}, pages 540--555.
  Springer, 2014.

\bibitem{unpaired}
Mrigank Rochan and Yang Wang.
\newblock Video summarization by learning from unpaired data.
\newblock In {\em Proceedings of the IEEE Conference on Computer Vision and
  Pattern Recognition}, pages 7902--7911, 2019.

\bibitem{rochan2018video}
Mrigank Rochan, Linwei Ye, and Yang Wang.
\newblock Video summarization using fully convolutional sequence networks.
\newblock In {\em Proceedings of the European Conference on Computer Vision
  (ECCV)}, pages 347--363, 2018.

\bibitem{sharghi2018improving}
Aidean Sharghi, Ali Borji, Chengtao Li, Tianbao Yang, and Boqing Gong.
\newblock Improving sequential determinantal point processes for supervised
  video summarization.
\newblock In {\em Proceedings of the European Conference on Computer Vision
  (ECCV)}, pages 517--533, 2018.

\bibitem{cvpr_2017}
Aidean Sharghi, Jacob~S Laurel, and Boqing Gong.
\newblock Query-focused video summarization: Dataset, evaluation, and a memory
  network based approach.
\newblock {\em Proceedings of the IEEE Conference on Computer Vision and
  Pattern Recognition}, 2017.

\bibitem{tvsum}
Yale Song, Jordi Vallmitjana, Amanda Stent, and Alejandro Jaimes.
\newblock Tvsum: Summarizing web videos using titles.
\newblock In {\em Proceedings of the IEEE Conference on Computer Vision and
  Pattern Recognition}, pages 5179--5187, 2015.

\bibitem{unsupervised_lstm_video_representation}
Nitish Srivastava, Elman Mansimov, and Ruslan Salakhudinov.
\newblock Unsupervised learning of video representations using lstms.
\newblock In {\em Proceedings of the IEEE Conference on Computer Vision and
  Pattern Recognition}, pages 843--852, 2015.

\bibitem{police_2013}
Jay Stanley.
\newblock Police body-mounted cameras: With right policies in place, a win for
  all.
\newblock {\em New York: ACLU}, 2013.

\bibitem{google_net}
Christian Szegedy, Wei Liu, Yangqing Jia, Pierre Sermanet, Scott Reed, Dragomir
  Anguelov, Dumitru Erhan, Vincent Vanhoucke, Andrew Rabinovich, et~al.
\newblock Going deeper with convolutions.
\newblock Cvpr, 2015.

\bibitem{cap_1}
Subhashini Venugopalan, Marcus Rohrbach, Jeffrey Donahue, Raymond Mooney,
  Trevor Darrell, and Kate Saenko.
\newblock Sequence to sequence-video to text.
\newblock In {\em Proceedings of the IEEE international conference on computer
  vision}, pages 4534--4542, 2015.

\bibitem{dpp_reference}
Anatoly Vershik.
\newblock {\em Asymptotic Combinatorics with Applications to Mathematical
  Physics: A European Mathematical Summer School held at the Euler Institute,
  St. Petersburg, Russia, July 9-20, 2001}.
\newblock Springer, 2003.

\bibitem{wang2009multi}
Feng Wang and Bernard Merialdo.
\newblock Multi-document video summarization.
\newblock In {\em 2009 IEEE International Conference on Multimedia and Expo},
  pages 1326--1329. IEEE, 2009.

\bibitem{video_retrieval}
Erkun Yang, Cheng Deng, Wei Liu, Xianglong Liu, Dacheng Tao, and Xinbo Gao.
\newblock Pairwise relationship guided deep hashing for cross-modal retrieval.
\newblock In {\em AAAI}, pages 1618--1625, 2017.

\bibitem{video_highlights_recurrent_autoencoders}
Huan Yang, Baoyuan Wang, Stephen Lin, David Wipf, Minyi Guo, and Baining Guo.
\newblock Unsupervised extraction of video highlights via robust recurrent
  auto-encoders.
\newblock In {\em Proceedings of the IEEE International Conference on Computer
  Vision}, pages 4633--4641, 2015.

\bibitem{yang2015unsupervised}
Huan Yang, Baoyuan Wang, Stephen Lin, David Wipf, Minyi Guo, and Baining Guo.
\newblock Unsupervised extraction of video highlights via robust recurrent
  auto-encoders.
\newblock In {\em Proceedings of the IEEE International Conference on Computer
  Vision}, pages 4633--4641, 2015.

\bibitem{rebuttal_5}
Ryo Yonetani, Kris~M Kitani, and Yoichi Sato.
\newblock Ego-surfing first-person videos.
\newblock In {\em Proceedings of the IEEE Conference on Computer Vision and
  Pattern Recognition}, pages 5445--5454, 2015.

\bibitem{gruman_summary_transfer}
Ke Zhang, Wei-Lun Chao, Fei Sha, and Kristen Grauman.
\newblock Summary transfer: Exemplar-based subset selection for video
  summarization.
\newblock In {\em Proceedings of the IEEE Conference on Computer Vision and
  Pattern Recognition}, pages 1059--1067, 2016.

\bibitem{zhang_lstm}
Ke Zhang, Wei-Lun Chao, Fei Sha, and Kristen Grauman.
\newblock Video summarization with long short-term memory.
\newblock In {\em European Conference on Computer Vision}, pages 766--782.
  Springer, 2016.

\bibitem{zhao2017hierarchical}
Bin Zhao, Xuelong Li, and Xiaoqiang Lu.
\newblock Hierarchical recurrent neural network for video summarization.
\newblock In {\em Proceedings of the 2017 ACM on Multimedia Conference}, pages
  863--871. ACM, 2017.

\bibitem{rl_diversity}
Kaiyang Zhou, Yu Qiao, and Tao Xiang.
\newblock Deep reinforcement learning for unsupervised video summarization with
  diversity-representativeness reward.
\newblock In {\em Thirty-Second AAAI Conference on Artificial Intelligence},
  2018.

\bibitem{rl_diversity_2}
Kaiyang Zhou, Tao Xiang, and Andrea Cavallaro.
\newblock Video summarisation by classification with deep reinforcement
  learning.
\newblock {\em The Thirty-Second AAAI Conference on Artificial Intelligence
  (AAAI-18)}, 2018.

\bibitem{action_1}
Wentao Zhu, Cuiling Lan, Junliang Xing, Wenjun Zeng, Yanghao Li, Li Shen,
  Xiaohui Xie, et~al.
\newblock Co-occurrence feature learning for skeleton based action recognition
  using regularized deep lstm networks.
\newblock In {\em AAAI}, volume~2, page~8, 2016.

\end{thebibliography}
